\documentclass{bmvc2k}

\usepackage{graphicx}
\usepackage{amsmath}
\usepackage{booktabs}
\usepackage{amssymb} 
\usepackage{subfigure}

\newcommand{\eg}{\emph{e.g.}} 
\newcommand{\ie}{\emph{i.e.}}

\title{Translational Concept Embedding for Generalized Compositional \\Zero-shot Learning}

\addauthor{He Huang}{hehuang@uic.edu}{1}
\addauthor{Wei Tang}{tangw@uic.edu}{1}
\addauthor{Jiawei Zhang}{jzhang@cs.fsu.edu}{2}
\addauthor{Philip S. Yu}{psyu@uic.edu}{1}

\addinstitution{
 Department of Computer Science\\
 University of Illinois at Chicago\\
 Chicago, USA
}
\addinstitution{
 Department of Computer Science\\
 Florida State University,\\
 Tallahassee, USA
}

\runninghead{Huang et al.}{Compositional Zero-shot Learning}

\def\eg{\emph{e.g}\bmvaOneDot}

\def\etal{\emph{et al}\bmvaOneDot}

\begin{document}

\maketitle

\begin{abstract}
Generalized compositional zero-shot learning means to learn composed concepts of attribute-object pairs in a zero-shot fashion, where a  model is trained on a set of seen concepts and tested on a combined set of seen and unseen concepts. This task is very challenging because of not only the gap between seen and unseen concepts but also the contextual dependency between attributes and objects. This paper introduces a new approach, termed \textit{translational concept embedding}, to solve these two difficulties in a unified framework.
It models the effect of applying an attribute to an object as adding a translational attribute feature to an object prototype. We explicitly take into account of the contextual dependency between attributes and objects by generating translational attribute features conditionally dependent on the object prototypes. Furthermore, we design a ratio variance constraint loss to promote the model's generalization ability on unseen concepts. It regularizes the distances between concepts by utilizing knowledge from their pretrained word embeddings. We evaluate the performance of our model under both the unbiased and biased concept classification tasks, and show that our model is able to achieve good balance in predicting unseen and seen concepts.
\end{abstract}


\section{Introduction}

Training machines to recognize visual concepts has been one of the central problems in artificial intelligence. A visual concept is usually composed of an \textit{attribute} and an \textit{object}, where the \textit{attribute} describes the state of an object or the modification applied to it. Recognizing visual concepts is more sophisticated than merely classifying object classes~\cite{he2016resnet,xian2018gbu}, and is beneficial to many applications such as scene understanding~\cite{huang2020addressing}, image captioning~\cite{yao2017boosting} and visual question answering~\cite{saqur2020multimodal}.

While humans can easily generalize from learned concepts such as \texttt{<red, apple>} and \texttt{<green, banana>} to an unseen concept \texttt{<green, apple>}, it remains a great challenge for machines to perform the same task. In this work, we study the problem of \textit{Generalized Compositional Zero-shot Learning} (GCZSL), where the model is trained on a set of \textit{seen} concepts of the form \texttt{<attribute, object>}, and is tested on a combined set of \textit{seen} and \textit{unseen} concepts. The \textit{seen} and \textit{unseen} concepts share the same set of \textit{attributes} and \textit{object} classes, but the compositions they form are disjoint in the \textit{seen} and \textit{unseen} sets. The objective of the task is to achieve as high accuracy as possible in recognizing \textit{unseen} concepts while maintaining good performance on \textit{seen} concepts.

Current approaches can be divided into three categories. The first approach decomposes the problem as two classification sub-problems: the model predicts the probability distributions of attributes and objects respectively~\cite{li2020symmetry}. The second approach adopts meta-learning techniques and treats the embedding of each \texttt{<attribute, object>} concept as task descriptions to directly measure the compatibility between the input image and a concept~\cite{purushwalkam2019tmn,wang2019tafe}. The third and most widely used approach is to project images and concepts as embeddings in a common latent space~\cite{nan2019pair_auto,nagarajan2018attributes,yang2020hidc,wei2019advfg}, and classification of an image is performed by finding the nearest neighbor concept in the latent space.


The task of GCZSL still remains challenging because it is difficult to compose visual representations of unseen concepts, especially when how an \textit{attribute} modifies the appearance is contextually dependent on the given \textit{object}.
Although current methods empirically work well on seen concepts, most of them fail to explicitly model the contextual dependency between attributes and objects and thus do not work as well on unseen concepts. 
In this work, we propose to model attributes as \textit{conditional translations} on \textit{objects} so that how an attribute modifies an object is explicitly conditioned on the object. 

The main motivation of modelling the effect of applying an attribute to an object as adding a \textit{translational attribute embedding} to the object prototype comes from unsupervised learning of word embeddings~\cite{mikolov2013distributed,pennington2014glove}. Word2Vec~\cite{mikolov2013distributed} finds that simple arithmetic operations (\eg, ``+'' and ``-'') reveal rich linear structures in the representation space. For example, the word vector operations Word2Vec(``King'') - Word2Vec(``Man'') + Word2Vec(``Woman'') result in a vector whose nearest neighbor is the Word2Vec vector of ``Queen''~\cite{mikolov2013distributed,radford2015dcgan}.
In order to better bridge the gap between \textit{seen} and \textit{unseen} concepts, based on the same intuition discussed above, we further propose a \textit{language prior constraint}. It regularizes the distances between different concepts to be proportional to the distances of their corresponding language embeddings~\cite{mikolov2013distributed,pennington2014glove}. This constraint is realized by designing a max-margin based \textit{ratio variance constraint} loss, which will be discussed in more details in Section~\ref{sec:rvc_loss}.

The main contributions of this paper are summarized as follows:
\begin{itemize}
    \item We propose a novel \textit{\textbf{T}ranslational \textbf{C}oncept \textbf{E}mbedding} (TCE) model. It explicitly takes into account the contextual dependency between attributes and objects via a translational attribute embedding conditioned on the object. By translating the object embedding according to the conditional attribute embedding, a composed concept representation can be learned more accurately.
    \item To further bridge the gap between seen and unseen concepts, we propose a novel \textit{ratio variance constraint} loss, which regularizes the distances between concept embeddings to be proportional to their distances in the word embedding space.
    \item We conduct extensive experiments with various metrics and show that our model works well on \textit{both seen and unseen concepts}, while previous works have stronger biases to either seen or unseen concepts. We also provide an open-source code base that implements many baseline methods to facilitate future research in this field.
\end{itemize}

\section{Related Work}

\textbf{Generalized Zero-shot Learning} (GZSL). The aim of GZSL is to train a classification model on a set of seen classes and test its performance on both seen and unseen classes~\cite{xian2018gbu,huang2019generative}. The goal is to achieve good performance on both seen and unseen classes. There are mainly two streams in the research of this problem, namely \emph{deterministic} and \emph{generative} approaches. The deterministic approaches~\cite{akata2013ALE,yang2018relation,liu2020hyperbolic} try to learn a visual-semantic mapping function that maps an image feature to the semantic feature of its corresponding class (or vice versa), where classification is performed by nearest-neighbor search.  The generative approaches~\cite{chen2018SPAEN,xian2019f,huang2019generative,huang2020multi,yu2020episode}, on the other hand, try to generate samples of unseen classes so that the zero-shot classification problem becomes a traditional classification problem with labeled data generated by the learned model. 
Although deterministic models are known for their simplicity and theoretical elegance, generative models~\cite{yu2020episode,keshari2020generalized} achieve superior results at the cost of trickier training techniques brought by generative adversarial nets (GANs)~\cite{goodfellow2014GAN} and variational auto-encoders (VAEs)~\cite{sohn2015cvae}. 

\textbf{Compositional Zero-shot Learning} (CZSL). CZSL is different from other zero-shot learning problems in that it focuses on classifying visual concepts that are composed of \textit{attributes} and \textit{objects}, and the large number of potential combinations as well as the contextual dependency between attributes and objects make this problem very challenging. The most naive approach is VisProd where we simply train two different classifiers for attributes and objects respectively, but this approach does not perform well due to the dependency between attributes and objects. AttrAsOp~\cite{nagarajan2018attributes} treats objects as feature vectors and attributes as matrices, and a concept feature vector is composed by multiplying an attribute matrix with an object feature vector. Although being effective, AttrAsOp still applies the same attribute matrix to different objects, which ignores the contextual dependency between them.  AdvFG~\cite{wei2019advfg} proposes to use multi-scale visual features and adversarial training to facilitate CZSL, while our method does not depend on those techniques that are not particularly related to the task. TaskMod~\cite{purushwalkam2019tmn} uses meta-learning and proposes to use modular gating networks to predict how relevant an image is to a given concept. As we show in the experiments, TaskMod is very biased towards seen classes and has inferior performance on unseen classes if the metric is unbiased accuracy. On the other hand, our proposed method achieves good balance on both seen and unseen classes. Nan \etal~\cite{nan2019pair_auto} propose to learn an auto-encoder that trains the model to reconstruct image features with the learned concept features.  HiDC~\cite{yang2020hidc} proposes to learn subspaces for attributes and objects, and compose the concept space using features sampled from the two subspaces. However, these two works also lack an explicit mechanism to capture the contextual dependency between attributes and objects.

\section{The Proposed Approach}
\label{sec:method}

\subsection{Notations and Definitions}
A visual concept $c \in \mathcal{C}$ is defined by a combination of an attribute $a\in\mathcal{A}$ and an object $o \in \mathcal{O}$: $c=(a,o)$, where $\mathcal{C}=\mathcal{A}\times\mathcal{O}$, $\mathcal{A}=\{a_1, a_2, ...,a_m\}$ and $\mathcal{O} = \{o_1, o_2,...,o_n\}$. The concept space $\mathcal{C}$ is split into two sets: a set of seen concepts $\mathcal{C}^s$ and a set of unseen concepts $\mathcal{C}^u$, where $\mathcal{C}^s \cup \mathcal{C}^u = \mathcal{C}$ and $\mathcal{C}^s \cap \mathcal{C}^u =\emptyset$. In generalized compositional zero-shot learning, the model is required to train on labeled images of seen concepts $\mathcal{C}^s$ and test on images of \textit{both seen and unseen concepts} $\mathcal{C}^s \cup \mathcal{C}^u$. Each attribute $a$ or object $o$ is associated with a semantic word embedding $\mathbf{e}_a$ or $\mathbf{e}_o$ initialized by Glove~\cite{pennington2014glove}. Given an image $\mathbf{I}_{c}$ of the concept $c=(a,o)$, we first extract an image feature using ResNet~\cite{he2016resnet} pretrained on ImageNet~\cite{ILSVRC15ImageNet}, and then pass it into a multi-layer perceptron (MLP) network to obtain its embedding $\mathbf{x}_{ao}$ in the common latent space. In the following sections, we will introduce how we obtain the concept embedding $\mathbf{\hat{x}}_{ao}$ of the concept $c=(a,o)$.

\begin{figure*}[t]
\begin{center}
  \includegraphics[width=0.95\linewidth]{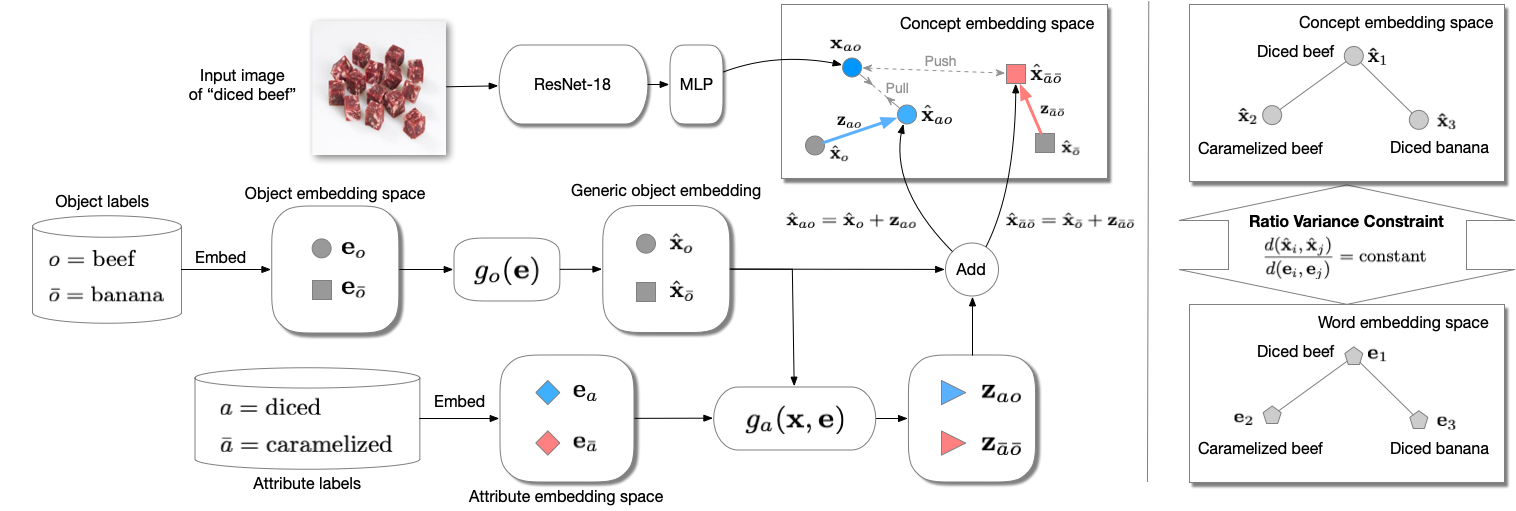}
\end{center}
  \caption{Our proposed TCE model that generates the attributes as translational embeddings applied on object prototypes. For each concept, we first generate an object prototype, then produce a translational attribute embedding conditioned on the object, and obtain the final concept embedding by adding the attribute embedding to the object prototype. We further design a \textit{ratio variance constraint} to regularize the distance between visual concepts by utilizing their pretrained word embeddings.  Detailed explanations in Section~\ref{sec:method}.} 
\label{fig:model}
\vspace{-0.5em}
\end{figure*}

\subsection{Generating Generic Object Prototypes}
As shown in Figure~\ref{fig:model}, in order to generate the concept embedding $\hat{\mathbf{x}}_{ao}$ for a concept $c=(a,o)$, we first embed the object $o$ in the object embedding space and obtain its representation $\mathbf{e}_o$, and then we pass it into a generation function $g_o(\cdot)$ to obtain a generic object prototype $\hat{\mathbf{x}}_{o}$ in the latent space $\mathcal{X}$:
\begin{align}
    \hat{\mathbf{x}}_{o} = g_o(\mathbf{e}_o), 
\end{align}

\subsubsection{Object Prototype Loss}
To ensure that the object prototype $\hat{\mathbf{x}}_{o}$ is representative of the object $o$, we train an object classifier $h^o_o(\cdot)$ on $\hat{\mathbf{x}}_{o}$ and minimize the cross-entropy loss:
\begin{align}
    \mathcal{L}^o_{cls} = \mathrm{CrossEntropy}(h_o^o(\hat{\mathbf{x}}_{o}), o),
\end{align}
where $h^o_o$ is implemented as an MLP network with softmax as its output activation function.

To better distinguish one object prototype from another, we randomly sample a negative object $\bar{o} \in \mathcal{O}$, where $\bar{o} \neq o$, and obtain its object prototype as $\hat{\mathbf{x}}_{\bar{o}}$. Then we apply a triplet max-margin loss~\cite{schroff2015triplet_loss}:
\begin{align}
\label{eq:obj_tri}
    \mathcal{L}^o_{tri} = \mathrm{max}(0, d(\mathbf{x}_{ao}, \hat{\mathbf{x}}_o) - d(\mathbf{x}_{ao}, \hat{\mathbf{x}}_{\bar{o}}) + m_o)),
\end{align}
where $d(\cdot,\cdot)$ is the euclidean distance function and $m_o$ is a hyper-parameter to control the margin. The overall object prototype loss $\mathcal{L}_{op}$ is thus defined as:
\begin{align}
    \mathcal{L}_{op} = \mathcal{L}^o_{cls} + \mathcal{L}^o_{tri}.
\end{align}

\subsection{Generating Concept Embeddings}
In order to apply the attribute $a$ to the generated object prototype $\hat{\mathbf{x}}_{o}$, we need to simultaneously take into account the object information as well as the attribute information since the same attribute may appear to be very different on various objects. In other words, the attribute representation of $a$ is contextually dependent on the object feature $\hat{\mathbf{x}}_{o}$. To this end, we design a neural network $g_a$ which takes the concatenation of the attribute embedding $\mathbf{e}_a$ and the generic object embedding $\hat{\mathbf{x}}_o$ as input, and generates the translational embedding $\mathbf{z}_{ao}$:
\begin{align}
    \mathbf{z}_{ao} = g_a(\hat{\mathbf{x}}_o, \mathbf{e}_a).
\end{align}
Then the final concept embedding $\hat{\mathbf{x}}_{ao}$ for $c=(a,o)$ can be obtained by translating the object prototype $\hat{\mathbf{x}}_o$ according to the translational attribute embedding $\mathbf{z}_{ao}$ as in Figure~\ref{fig:model}:
\begin{align}
    \hat{\mathbf{x}}_{ao} = \hat{\mathbf{x}}_o + \mathbf{z}_{ao}.
\end{align}

To take advantage of contrasting learning as in object prototypes generation, for each training image $\mathbf{I}_c$ of concept $c=(a, o) \in \mathcal{C}^s$, we randomly sample a negative concept $\bar{c} = (\bar{a},\bar{o})$, where $\bar{c} \neq c$ and $\bar{c} \in \mathcal{C}^s$. The concept embedding $\hat{\mathbf{x}}_{\bar{a}\bar{o}}$ of the negative concept $\bar{c}$ can be obtained through the same process as generating $\hat{\mathbf{x}}_{ao}$.

\subsubsection{Concept Embedding Loss}
The concept embedding $\hat{\mathbf{x}}_{ao}$ is desired to be representative of the concept $c=(a,o)$, thus we use two classifiers $h^c_a(\cdot)$ and $h^c_o(\cdot)$ for attributes and objects respectively and minimize the CrossEntropy loss:
\begin{align}
    \mathcal{L}_{cls} =\ & \mathrm{CrossEntropy}(h^c_a(\hat{\mathbf{x}}_{ao}), a) \\ \notag
    & + \mathrm{CrossEntropy}(h^c_o(\hat{\mathbf{x}}_{ao}), o),
\end{align}
where $h^c_a$ and $h^c_o$ are implemented as MLPs with softmax as the last activation.

Similar to the triplet margin loss we use in Equation~\ref{eq:obj_tri}, we enforce a minimal margin between the distance of the positive pair $(\mathbf{x}_{ao}, \hat{\mathbf{x}}_{{a}{o}})$ and the negative pair  $(\mathbf{x}_{ao}, \hat{\mathbf{x}}_{\bar{a}\bar{o}})$:
\begin{align}
    \mathcal{L}_{tri} = \mathrm{max}(0, d(\mathbf{x}_{ao}, \hat{\mathbf{x}}_{ao}) - d(\mathbf{x}_{ao}, \hat{\mathbf{x}}_{\bar{a}\bar{o}}) + m_c)),
\end{align}
where $d(\cdot,\cdot)$ is the euclidean distance function and $m_c$ is a hyper-parameter margin.

Since the triplet margin loss only enforces a minimum margin between $d(\mathbf{x}_{ao}, \hat{\mathbf{x}}_{{a}{o}})$ and $d(\mathbf{x}_{ao}, \hat{\mathbf{x}}_{\bar{a}\bar{o}})$, minimizing the triplet margin loss is similar to enforcing a minimum distance between the positive $\hat{\mathbf{x}}_{{a}{o}}$ and negative $\hat{\mathbf{x}}_{\bar{a}\bar{o}}$ concept embeddings, and thus it is possible that 
after training $d(\mathbf{x}_{ao}, \hat{\mathbf{x}}_{{a}{o}}) \gg d(\hat{\mathbf{x}}_{\bar{a}\bar{o}}, \hat{\mathbf{x}}_{{a}{o}}) \ge m_c$. In other words, 
the learned concept embeddings may be clustered to some degree, depending on the choice of margin $m_c$. Such a situation may hurt the model's performance because we would like a concept embedding to be close to the center of the image embeddings of the same concept while being far away from the other concept embeddings. To fix this problem, we add a reconstruction loss which pulls the concept embedding $\hat{\mathbf{x}}_{ao}$ towards the embedding of
input image $\mathbf{x}_{ao}$:
\begin{align}
    \mathcal{L}_{rec} = d(\mathbf{x}_{ao}, \hat{\mathbf{x}}_{{a}{o}})
\end{align}

The overall training loss for learning concept embeddings is defined as a weighted combination of the above three loss functions:
\begin{align}
    \mathcal{L}_{conc} = \lambda_{cls}\mathcal{L}_{cls}  + \lambda_{tri}\mathcal{L}_{tri} + \lambda_{rec}\mathcal{L}_{rec}, 
\end{align}
where $\lambda_{cls}$, $\lambda_{tri}$ and $\lambda_{rec}$ are hyper-parameters.

\subsection{Ratio Variance Constraint}
\label{sec:rvc_loss}

To better bridge the gap between seen and unseen concepts, here we propose another loss function named \textit{ratio variance constraint}. The motivation comes from previous study on image synthesis with generative adversarial networks (GANs). DCGAN~\cite{radford2015dcgan} finds that there is a strong correlation between the linear operation of image latent features and the semantic meaning of the resulting output images. For example, the latent feature of an image of \texttt{<smiling, woman>}, when subtracted with the latent feature of an image of \texttt{<woman>} and added with the latent feature of an image of \texttt{<man>}, will produce an image of the content \texttt{<smiling, man>}, \ie, $G$(vector(``smiling woman'') - vector(``woman'') + vector(``man'')) = $G$(vector(``smiling man'')) where $G(\cdot)$ is the learned generator. In other words, the linear structure space discovered in unsupervised word representation learning~\cite{mikolov2013distributed} also exists in the image latent feature space, and there is a strong correlation between word embeddings and image latent features. 

In order to utilize this image-text correlation to bridge the gap between seen and unseen concepts, we propose a \textit{ratio variance constraint} loss, which regularizes the distances between learned concept embeddings, as illustrated in the right part of Figure~\ref{fig:model}. Given the latent features $\hat{\mathbf{x}}_{c_i}, \hat{\mathbf{x}}_{c_j}$ of two concepts $c_i, c_j$, we would like their distance to be proportional to the distance between their semantic word embeddings $\mathbf{e}_{c_i}, \mathbf{e}_{c_j}$:
\begin{align}
    \frac{{d}(\hat{\mathbf{x}}_{c_i}, \hat{\mathbf{x}}_{c_j})}{{d}(\mathbf{e}_{c_i}, \mathbf{e}_{c_j})} = \mathrm{constant}, \forall c_i,c_j \in \mathcal{C}\ \mathrm{and}\ c_i \neq c_j, 
\label{eq:ratio}
\end{align}
where $d(\cdot, \cdot)$ represents the euclidean distance function, and the semantic embedding $\mathbf{e}_c$ of a concept $c=(a,o)$ is obtained by adding $\mathbf{e}_a$ and $\mathbf{e}_o$ of its corresponding attribute $a$ and object $o$.
The above constraint is very strong, which creates great challenges in model optimization. To facilitate model learning, we propose to relax it and introduce the following variance based loss term approximation. Formally, to satisfy Equation~\ref{eq:ratio}, we randomly sample pairs of $(c_i, c_j)$ and minimize the variance. In addition, we allow the variance to be larger than 0 but smaller than a margin $m_r$:

\begin{align}
    \mathcal{L}_{\mathrm{rvc}} = \max(0, var(\{\frac{{d}(\hat{\mathbf{x}}_{c_i}, \hat{\mathbf{x}}_{c_j})}{{d}(\mathbf{e}_{c_i}, \mathbf{e}_{c_j})}, \forall c_i,c_j \in \mathcal{C}\ \mathrm{and}\ c_i \neq c_j\}) - m_r),
\end{align}
where $var(\cdot)$ is the variance function. In practice, we randomly sample 100 pairs of $(c_i, c_j)$ to calculate the loss.

\subsection{Training and Inference}

The overall loss function to be optimized is defined as:
\begin{align}
    \mathcal{L} = \mathcal{L}_{conc} + \lambda_{op}\mathcal{L}_{op} + \lambda_{rvc}\mathcal{L}_{rvc},
\end{align}
where $\lambda_{op}$ and $\lambda_{rvc}$ are hyper-parameters.
During inference, we enumerate all concepts $c=(a,o) \in \mathcal{C}^s \cup \mathcal{C}^u$, and generate its corresponding concept embedding $\hat{\mathbf{x}}_{c}$. Given a testing image, we obtain its embedding $\mathbf{x}$ in the learned common latent space, and assign it to the concept which has the shortest euclidean distance to it.



\section{Experiments}
\vspace{-1em}
\textbf{Datasets:} 
We consider two different datasets \textbf{MIT-States}~\cite{isola2015mitstates} and \textbf{UT-Zappos}~\cite{yu2014zappos}. MIT-States has a very diverse set of 245 objects (\eg,\ ``beef'' and ``tower'') and 115 attributes (\eg,\ ``diced'' and ``foggy''), with over 53K images. UT-Zappos consists of different types of shoes (\eg, ``rubber sneaker'' and ``leather sandal''), with 12 object classes, 16 attribute classes and a total about 33K images. Since the dataset split used in some previous work~\cite{yang2020hidc,nagarajan2018attributes} only tests on images of unseen concepts, in order to evaluate under the GCZSL setting, we follow the data split used in \cite{purushwalkam2019tmn}, which contains images of seen and unseen concepts for both validation and testing. 

\textbf{Metrics:} We evaluate our model's performance under a various set of metrics. (1) \textbf{closed unseen} accuracy, where each image $\mathbf{I}_c\ (c \in \mathcal{C}^u)$ is only classified among $\mathcal{C}^u$. (2) \textbf{open unseen} accuracy, where each image $\mathbf{I}_c\ (c \in \mathcal{C}^u)$ is classified among $\mathcal{C}^u\cup \mathcal{C}^s$. (3) \textbf{open seen} accuracy, where each image $\mathbf{I}_c\ (c \in \mathcal{C}^s)$ is classified among $\mathcal{C}^u\cup \mathcal{C}^s$. (4) \textbf{unseen HM} is the \emph{harmonic mean} of \emph{closed unseen} and \emph{open unseen}. (5) \textbf{all HM} is the \emph{harmonic mean} of \emph{open unseen} and \emph{open seen}, which is our major metric. (6) \textbf{AUC}~\cite{purushwalkam2019tmn,chao2016empirical} is calculated by first adding various bias weights to the \textit{unseen concepts} to make the prediction bias towards seen or unseen concepts, and then calculate the area under the curve of open seen-unseen accuracy. (7) Accuracy on separate attribute (\textbf{attr acc}) and object (\textbf{obj acc}) classification.

\textbf{Baselines:}
We compare our model with several baselines. \textbf{VisProd} uses individual MLPs in classifying attributes and objects, and calculates the probability of a concept $c=(a,o)$ as $P(c) = P(a)P(o)$. \textbf{LabelEmbed+}~\cite{nagarajan2018attributes} concatenates the pretrained word embeddings of attributes and objects and uses a single MLP network to generate the concept embedding. \textbf{AttrAsOp}~\cite{nagarajan2018attributes} learns a matrix for each attribute and a vector for each object, and composes the concept embedding by matrix-vector multiplication. \textbf{TaskMod}~\cite{purushwalkam2019tmn} learns the correlation between an image and a concept with modular gating networks. \textbf{AdvFG}~\cite{wei2019advfg} uses semi-negative samples in triplet margin loss and trains with an adversarial loss. \textbf{PairAE}~\cite{nan2019pair_auto} uses an auto-encoder to enforce that the concept embeddings can reconstruct the ResNet~\cite{he2016resnet} image features. \textbf{HiDC}~\cite{yang2020hidc} proposes to learn a subspace for attributes and objects respectively and then learn to combine them into a concept space. 

\textbf{Implementation Details:}
All compared models use ResNet-18~\cite{he2016resnet} as image feature extractor. For AttrAsOp and TaskMod, we use the code released by their authors\footnote{https://github.com/Tushar-N/attributes-as-operators}\footnote{https://github.com/facebookresearch/taskmodularnets}. For all other baselines without public code, we implemented them using the hyper-parameters in the papers, and did our best to tune those that are not provided by the authors. 
Our code, including all compared methods and configurations, is publicly available to benefit future research. See more details in the Appendix \ref{app:imp}.

\begin{table*}[!ht]
\centering
\caption{Comparison with baseline methods on unbiased Generalized Compositional Zero-Shot Learning (GZSL).}
\label{tab:results}
\resizebox{\textwidth}{!}{
\begin{tabular}{l| rrrrr| rrrrr}\hline
 & \multicolumn{5}{c|}{MIT-States} & \multicolumn{5}{c}{UT-Zappos} \\ 
Model & \multicolumn{1}{l}{closed unseen} & \multicolumn{1}{l}{open unseen} & \multicolumn{1}{l}{open seen} & \multicolumn{1}{l}{unseen HM} & \multicolumn{1}{l|}{all HM} & \multicolumn{1}{l}{closed unseen} & \multicolumn{1}{l}{open unseen} & \multicolumn{1}{l}{open seen} & \multicolumn{1}{l}{unseen HM} & \multicolumn{1}{l}{all HM} \\ \hline\hline
VisProd & 3.17 & 3.07 & \textbf{19.12} & 3.11 & 5.29 & 7.24 & 7.22 & 55.71 & 7.22 & 12.78 \\
LabelEmbed+~\cite{nagarajan2018attributes} & 5.56 & 5.32 & 11.48 & 5.43 & 7.27 & 15.61 & 15.49 & 49.46 & 15.54 & 23.59 \\
AttrAsOp~\cite{nagarajan2018attributes} & 12.45 & 10.77 & 2.54 & 11.54 & 4.11 & 26.81 & 25.85 & 33.04 & 26.32 & 29.01 \\
TaskMod~\cite{purushwalkam2019tmn} & 3.94 & 3.69 & 15.45 & 3.81 & 5.95 & 21.65 & 15.97 & 32.14 & 18.38 & 21.33 \\
AdvFG~\cite{wei2019advfg} & 8.68 & 7.32 & 5.71 & 7.94 & 6.41 & 31.15 & 29.84 & 33.43 & 30.48 & 31.53 \\
PairAE~\cite{nan2019pair_auto} & 4.56 & 4.24 & 14.39 & 4.39 & 6.55 & 19.72 & 19.14 & 32.55 & 19.42 & 24.10 \\
HiDC~\cite{yang2020hidc} & 8.20 & 6.75 & 9.45 & 7.40 & 7.87 & 17.92 & 17.34 & \textbf{58.16} & 17.62 & 26.71 \\ \hline
TCE (ours) & \textbf{12.46} & \textbf{11.55} & 8.27 & \textbf{11.99} & \textbf{9.64} & \textbf{31.36} & \textbf{30.68} & {42.52} & \textbf{31.01} & \textbf{35.64}\\ \hline
\end{tabular}
}
\vspace{-1.5em}
\end{table*}

\subsection{Results}

\subsubsection{Unbiased Concept Classification}
The main results are in Table~\ref{tab:results}. The proposed TCE model achieves the highest \textit{all HM} on both datasets, with 1.77\% and 4.11\% absolute increase over the second best on MIT-States and UT-Zappos respectively. On the MIT-States dataset, our \textit{closed unseen} accuracy is close to AttrAsOp~\cite{nagarajan2018attributes}, but our \textit{open unseen} and \textit{open seen} scores are much higher than that of AttrAsOp. For models that have higher \textit{open seen} scores than ours on MIT-States and UT-Zappos, such as TaskMod~\cite{purushwalkam2019tmn}, PairAE~\cite{nan2019pair_auto} and HiDC~\cite{yang2020hidc}, we have much higher \textit{unseen} scores than them. Overall, our model is able to compose representative embeddings for unseen concepts while preserving good performance on seen classes. It can also be noted that all methods achieve much higher scores on the UT-Zappos than the MIT-States dataset, which is because of the fact that UT-Zappos has a much smaller label space than MIT-States, and that UT-Zappos only contains different shoes, while MIT-States contains objects of large variety which makes the task even harder.
Our reproduced results for \textit{AttrAsOp} are very close to the scores in its original paper, since its code contains all hyper-parameters needed to train the model. 
For \textit{TaskMod}, training with Softmax introduces a strong bias towards seen classes, which makes \textit{TaskMod} perform worse on unseen concepts under the unbiased setting, especially on the MIT-States dataset. 
For the other baselines (\ie, AdvFG, PairAE, HiDC), we were not able to obtain the exact set of hyper-parameters required to reproduce their results, but we did our best to tune their performance. Overall, they have relatively better \textit{all HM} compared to AttrAsOp and TaskMod on MIT-States, and they are comparable to or better than other baselines on UT-Zappos.

\subsubsection{Biased Concept Classification}

In this setting, we add a set of weights to the unseen concepts to bias the prediction towards either \textit{seen} or \textit{unseen} concepts, as done in~\cite{purushwalkam2019tmn,chao2016empirical}. For each prediction, we first obtain the maximum score among all concepts, denoted as $s_{max}$, then we divide the interval $[-s_{max}, s_{max}]$ evenly into 100 bins and use them as biases added to unseen concepts. The results are extrapolated as curves, and we calculate the \textit{area under the curve} (AUC) (divided by 100) and show them in Table~\ref{tab:results2}. 
We can see that our model is able to achieve the highest AUC on both MIT-States and UT-Zappos. Combined with the results in Table~\ref{tab:results}, we can see that our model is able to achieve good performance on both biased and unbiased metrics.
For baselines that have strong bias towards seen classes on MIT-States, such as VisProd and TaskMod, they achieve good AUC under the biased setting, even though they have lower unseen accuracy in Table~\ref{tab:results}. 
Overall, the AUC of different methods are close to each other, which shows that the model's internal bias towards seen or unseen classes plays an important role in the performance difference under the unbiased setting. In practice, it is hard to choose an appropriate bias to use, and thus the unbiased setting is more applicable in real-world situations, since we want the model to perform well on both seen and unseen classes.

\subsubsection{Separate Attribute and Object Classification}

We show the separate classification accuracy for attribute and object classification in Table~\ref{tab:results2}. We can see that our method is able to achieve the highest accuracy for both attribute and object classification on the MIT-States dataset, while on the UT-Zappos dataset we achieve the best object accuracy and the second best attribute accuracy that is competitive to the best HiDC. Although HiDC are better at predicting object attributes than ours on UT-Zappos, our method perform much better on attribute-object concept classification, as shown in Table~\ref{tab:results}.

\begin{table}[]
\centering
\caption{Comparing AUC on biased concept classification, and individual attribute/object classification accuracy.}
\hspace{0.5em}
\label{tab:results2}
\resizebox{0.55\textwidth}{!}{
\begin{tabular}{|l|c|c|c|c|c|c|}
\hline
 & \multicolumn{3}{c|}{MIT-States} & \multicolumn{3}{c|}{UT-Zappos} \\ \hline
Model & AUC & attr acc & obj acc & AUC & attr acc & obj acc \\ \hline
VisProd & 1.96 & 18.83 & 25.21 & 28.92 & 41.48 & 66.51 \\ \hline
LabelEmbed+ & 1.91 & 17.91 & 22.31 & 26.29 & 40.18 & 70.89 \\ \hline
AttrAsOp & 1.93 & 18.11 & 22.33 & 27.04 & 36.61 & 68.36 \\ \hline
TaskMod & 1.17 & 17.13 & 24.29 & 26.54 & 44.44 & 64.44 \\ \hline
AdvFG & 2.18 & 17.95 & 21.11 & 27.82 & 40.97 & 68.18 \\ \hline
PairAE & 1.96 & 17.36 & 26.12 & 28.48 & 39.37 & 66.67 \\ \hline
HiDC & 2.00 & 16.97 & 22.54 & {30.79} & \textbf{45.26} & 70.01 \\ \hline
TCE & \textbf{2.23} & \textbf{19.13} & \textbf{26.35} & \textbf{31.95} & 44.73 & \textbf{71.18} \\ \hline
\end{tabular}
}
\end{table}

\begin{table*}[]
\centering
\caption{Ablation study on loss components.}
\label{tab:ablation}
\resizebox{\textwidth}{!}{
\begin{tabular}{l|rrrrr|rrrrr}\hline
 & \multicolumn{5}{c|}{MIT-States} & \multicolumn{5}{c}{UT-Zappos} \\
TCE variant & \multicolumn{1}{l}{attr acc} & \multicolumn{1}{l}{obj acc} & \multicolumn{1}{l}{open unseen} & \multicolumn{1}{l}{open seen} & \multicolumn{1}{l|}{all HM} & \multicolumn{1}{l}{attr acc} & \multicolumn{1}{l}{obj acc} & \multicolumn{1}{l}{open unseen} & \multicolumn{1}{l}{open seen} & \multicolumn{1}{l}{all HM} \\ \hline \hline
(1) $+\mathcal{L}_{tri}$ & 18.62 & 24.91 & 6.21 & 12.66 & 8.33 & 42.27 & 69.80 & 17.13 & 50.34 & 25.56 \\
(2) $+\mathcal{L}_{tri}+\mathcal{L}_{cls}$ & 17.40 & 21.27 & 9.90 & 2.15 & 3.53 & 45.16 & 68.57 & 26.22 & 44.18 & 32.90 \\
(3)$+\mathcal{L}_{tri}+\mathcal{L}_{cls}+\mathcal{L}_{rec}$ & 18.52 & 26.11 & 8.93 & 9.11 & 9.01 & 43.17 & 68.61 & 21.16 & 47.99 & 29.37 \\
(4) $+\mathcal{L}_{tri}+\mathcal{L}_{cls}+\mathcal{L}_{rec}+\mathcal{L}_{op}$ & 18.67 & 26.41 & 9.48 & 8.81 & 9.13 & 43.76 & 70.41 & 24.59 & 46.91 & 32.26 \\
(5) $+\mathcal{L}_{tri}+\mathcal{L}_{rec}+\mathcal{L}_{op}+\mathcal{L}_{rvc}$ & 15.52 & 25.14 & 4.27 & 13.72 & 6.51 & 39.77 & 66.27 & 6.77 & 56.21 & 12.08 \\
(6) $+\mathcal{L}_{tri}+\mathcal{L}_{cls}+\mathcal{L}_{rec}+\mathcal{L}_{op}+\mathcal{L}_{rvc}$ & 19.13 & 26.35 & 11.55 & 8.27 & 9.64 & 44.73 & 70.35 & 30.68 & 42.52 & 35.64 \\ \hline  
\end{tabular}
}
\end{table*}

\subsection{Ablation Study}

In this section we study the effects of each loss term in our model, and the results are shown in Table~\ref{tab:ablation}. (1) We can see that the basic $\mathcal{L}_{tri}$ loss strongly biases towards the seen concepts, although it achieves good harmonic mean between seen and unseen concepts. (2) Adding $\mathcal{L}_{cls}$ alone helps the prediction on unseen concepts but degrades the performance on seen classes. (3) Adding $\mathcal{L}_{rec}$ can enforce the concept embeddings to be close to the image embeddings of the same concept, which improves the accuracy on seen concepts. (4) $\mathcal{L}_{op}$ improves the object and attribute classification accuracy as well as the performance on unseen concepts, which shows the effectiveness of first generating good object prototypes before generating concept embeddings. (5) From (2) we notice that $\mathcal{L}_{cls}$ has a negative effect on many metrics, so here we study the effect of the whole model without $\mathcal{L}_{cls}$. We can see that the model has a even stronger bias towards seen concepts than all other variants, which shows that $\mathcal{L}_{cls}$ is still a necessary component to help the model generalize to unseen concepts. (6) By adding $\mathcal{L}_{rvc}$ to all previous loss components, we further improve the performance on unseen concepts by regularizing distances between concepts and bridging the gap between seen and unseen concepts.This shows that the difference of two concepts in word embedding space is a good indicator of the relation of the two concepts in the image feature space. Overall, we can still see that there is a trade-off between the performance of seen and unseen classes. On one hand, $\mathcal{L}_{tri}$ and $\mathcal{L}_{rec}$ both try to minimize the distance between two training samples of the same seen class, and no information of unseen classes is involved in the two loss terms, thus they tend to bias towards the seen classes. On the other hand, $\mathcal{L}_{cls}$ tries to classify the individual attributes and objects correctly, while ignoring the compositions of attributions and objects, thus it does not work well on concept classification when applied alone. However, since the seen and unseen concepts share the same set of attributes and objects, $\mathcal{L}_{cls}$ can help generalize the knowledge learned from seen classes to unseen classes. 

\begin{figure}[!t]
\centering
\subfigure[]{
\begin{minipage}[l]{0.45\linewidth}
\centering
\includegraphics[width=0.9\textwidth]{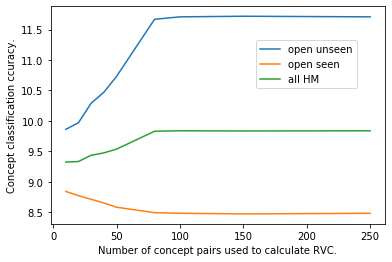}
\end{minipage}
\label{fig:npairs}
}
\subfigure[]{
\begin{minipage}[l]{0.45\linewidth}
\centering
\includegraphics[width=0.9\textwidth]{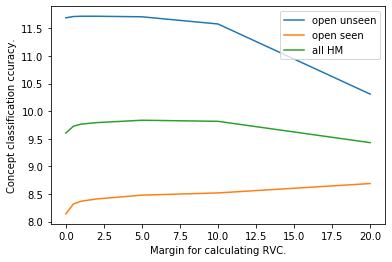}
\end{minipage}
\label{fig:margin}
}
\caption{Ablation study on the \emph{ratio variance constraint} (RVC) loss evaluated on the validation set: (a) Increasing the number of concept pairs used in RVC helps improving the unseen accuracy and h-mean and saturates around 100. (b) Using a smaller margin $m_r$ in RVC allows a tighter alignment between the concept embedding space and semantic word embedding space, thus helping the model to generalize to unseen concepts.}
\label{fig:rvc}
\vspace{-0.5em}
\end{figure}

We further study the effect of two hyper-parameters used in calculating \emph{ration variance constraint} (RVC). We tune the hyper-parameters using validation data and show the performance on validation set, as illustrated in Figure~\ref{fig:rvc}.  Figure~\ref{fig:npairs} shows the effect of varying the number of concept pairs used in calculating RVC. We can see that using a larger number of pairs generally improves the performance on unseen classes, since it helps to estimate a more accurate variance. The performance saturates when using about 100 pairs. Although RVC has a $\sim0.5\%$ accuracy decrease on seen concepts, its gain on unseen concepts ($\sim1.5\%$) is larger than the decrease.
Figure~\ref{fig:margin} shows the effect of varying the margin $m_r$ in RVC. Using a smaller margin has better unseen and h-mean accuracy, because a smaller margin enforces a tighter alignment between the concept embedding space and the semantic embedding space, which makes it easier to generalize from seen to unseen concepts. When $m_r=0$, the model's performance on seen concepts has a sharp drop, which shows that using a hard alignment is not ideal, and a soft alignment ($m_r>0$) allows a better balance between seen and unseen concepts. When increasing the margin $m_r$ to larger numbers ($m_r>10$), the RVC applies a very week constraint on the alignment between the two spaces. In this case, the model's performance on unseen concepts is decreased when compared with smaller $m_r$, because the model learns much less knowledge from the concepts' relations in the semantic embedding space.

\begin{figure*}[!hbt]
\begin{center}
   \includegraphics[width=0.99\linewidth]{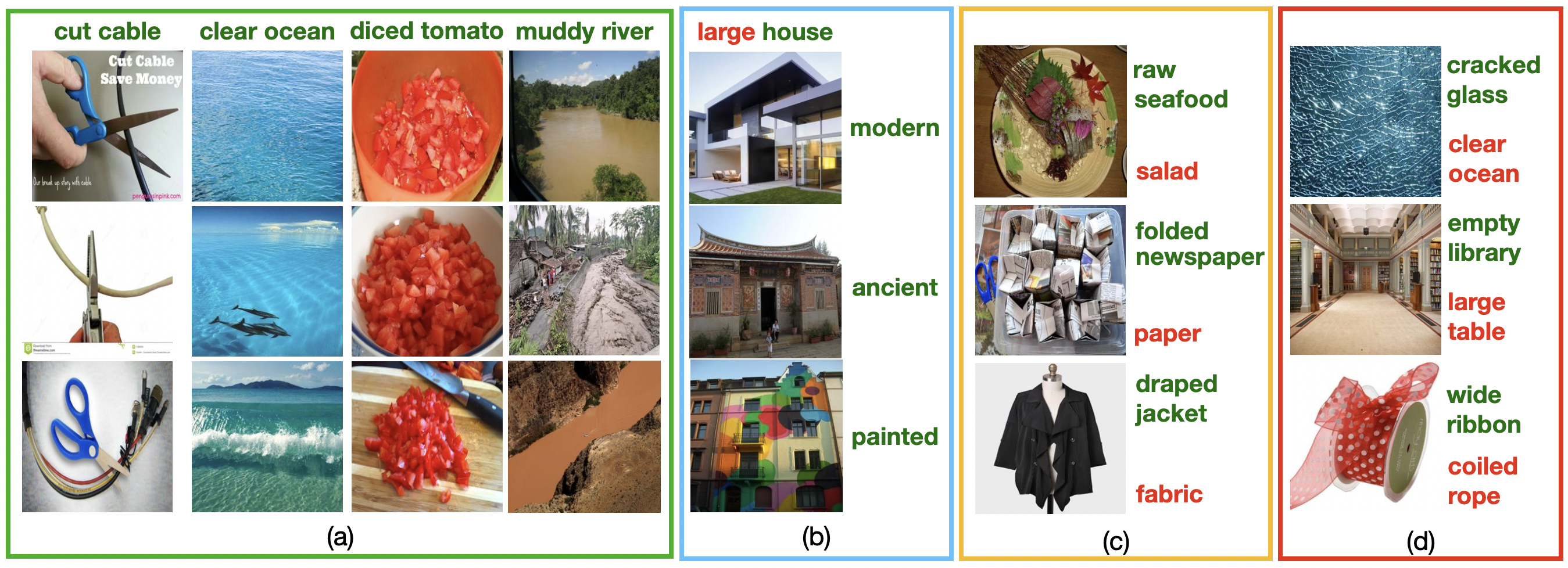}
\end{center}
   \caption{Qualitative results of our method on MIT-States (green color text for groundtruth, red for incorrect prediction): (a) both attributes and objects are predicted correctly. (b) correct objects but incorrect attributes. (c) correct attributes but incorrect objects. (d) both attributes and objects are predicted incorrectly.
   } 
\label{fig:vis}
\end{figure*}

\subsection{Qualitative Results}
In Figure~\ref{fig:vis}, we show some qualitative results of our method on MIT-States. We can see that our model works pretty well on concepts that are well defined and without ambiguity (\eg, Figure~\ref{fig:vis}(a)). Figure~\ref{fig:vis}(b) shows that our model fails on some cases where the objects can be described by more than one of the attributes. For example, all three houses in Figure~\ref{fig:vis}(b) could be considered as ``large'', in addition to their actual labelled attributes. Figure~\ref{fig:vis}(c) shows that our model could fail in cases where the objects are semantically non-exclusive. For example, ``seafood" could also appear in ``salad", ``newspaper" is similar to ``paper" with words, and ``jacket" is made of ``fabric". Figure~\ref{fig:vis} shows the cases where our model fail to predict either attributes or objects. In these cases, the images are similar to the incorrectly predicted concepts to some degree (\eg, local patterns). To solve this problem, it would be better to use the image feature maps instead of global-averaged features, so that the model can explore and attend to different locations when classifying the concepts. We leave this to future research.

\section{Conclusion}
In this paper, we study the GCZSL problem by proposing a \textit{translational concept embedding} (TCE) model. We propose to model the effect of applying an attribute to an object as adding a residual attribute feature to a generic object prototype. We design a set of loss functions to train the TCE model, and especially we propose a \textit{ratio variance constraint} to bridge the gap between seen and unseen concepts. 
Experiments and ablation study prove the effectiveness of our model under both the biased and unbiased concept classification settings, and we provide a publicly available code base with many baselines to benefit future research.

\bibliography{reference}

\appendix
\section{Appendix}
\subsection{Extended Related Work}
\cite{naeem2021learning} propose to use graph convoluational  networks (GCN) to learn concept embeddings. In the graph, each attribute, object and concept represents an individual node, and there is an edge between an attribute/object node and a concept node if the attribute/object is used in the concept. Here they study the compositional zero-shot learning (CZSL) problem under a \emph{transductive} setting, since they use the unseen attribute-object concepts to build the graph used in both training and testing, while our model dose not have access to the test concepts. Additionally, they finetune the image feature extractor along with training the GCN, while previous methods all used fixed image feature extractor.

\cite{mancini2021open} study the open-world setting of CZSL, where there are infeasible concepts that don't belong to either seen or unseen concepts during classification. To solve this problem, they propose to learn a feasibility score for each constructed attribute-object concept, and then use this score to calibrate the softmax logits for classification.

\cite{atzmon2020causal} propose a generative model to learn concept embeddings. Our method is different from theirs in that we use a deterministic approach which is easier to train than generative approaches, and we utilize the correlation between concepts in the semantic word embedding space.

\subsection{Implementation Details}
\label{app:imp}
Given the fact that our experiment setting is different from all previous works, and that most baselines do not have open-source code, we implement most baselines and tune their hyper-parameters with our best effort. Here we provide details of how we re-implement each baseline. Since \textbf{AttrAsOp}~\cite{nagarajan2018attributes} and \textbf{TaskMod}~\cite{purushwalkam2019tmn} have their code and hyper-parameters available, we just use the code from the users\footnote{https://github.com/Tushar-N/attributes-as-operators}\footnote{https://github.com/facebookresearch/taskmodularnets}. Our codebase is implemented using PyTorch-v1.4 and will be publicly available after the paper being published, including the hyper-parameters we use. All experiments are run on a NVIDIA  Titan-RTX GPU with 24GB memory on a Ubuntu 16.04 system. In the following sections, we explain the details of how we implement each baseline and the hyper-parameters we use. The same set of hyper-parameters are used for both datasets unless specified otherwise. All feature extractors are ResNet-18~\cite{he2016resnet} without finetuning, and the 512-dimensional feature after the last global average pooling layer is used as the image feature. All baselines are trained with a maximum epoch of 1200 and batch size 512 unless specified otherwise. All baselines share the same codebase, which means the training and evaluation pipeline are the same across different methods to make the results more directly comparable. 

\subsubsection{VisProd} We use a two-layer multi-layer perceptron (MLP) network, where the hidden dimension is 512 with ReLU activation for attribute and object classification respectively. The optimizer is Adam with learning rate of $1e-4$ and weight decay $5e-5$.

\subsubsection{LabelEmbed+} This model is implemented as a single 2-layer MLP network, where the input (\ie, concatenation of attribute and object embeddings) and the hidden dimension are 600, and the output dimension (\ie, concept embedding dimension) is 300. The attribute and object embeddings are initialized with 300-dimensional pretrained GloVe~\cite{pennington2014glove} embeddings and are finetuned during training. The optimizer is Adam with learning rate of $1e-4$ and weight decay $5e-5$. Margin of the triplet margin loss is 0.5 for all experiments.

\subsubsection{AdvFG} In the original paper~\cite{wei2019advfg}, the authors extract a 960-dimensional visual feature using ResNet-18. However, the paper does not provide what layers of ResNet-18 to use for the extracting the multi-scale features. In order to make this model directly comparable to others, we use the same 512-dimensional ResNet-18 feature as other baselines. The paper also does not provide details on how to obtain the ``960-dimensional linguistic feature vector for both attribute and object with word embeddings'', so we use the same 300-dimensional GloVe embeddings as other baselines.  

For the network configurations, since the paper dose not provide the information, we define the number of MLP layers to be equal to the number of layers shown in the Figure 2 of the paper. Specifically, the generator and discriminator are implemented as 2-layer MLP with 300 and 128 hidden dimension respectively. The other networks are implemented as single-layer MLP. Activation functions are all ReLU except for the attribute/object classifier (Softmax) and the last layer of the discriminator (Sigmoid). The concept embedding dimension is set to 300 as it is not specified in the paper.

All other hyper-parameters are set to the values in the paper~\cite{wei2019advfg}. 

\subsubsection{PairAE}
For PairAE~\cite{nan2019pair_auto}, we use the exact same set of hyper-parameters provided in the paper. Since the paper does not specify the kind of activation function used in each MLP, we set it to ReLU. The learning rate, learning rate decay factor and dropout are all set to the values in the paper.

\subsubsection{HiDC}
This paper is missing key implementation details, and we could not find its code. We tried emailing the authors but they said they lost the original version of the code and needed to re-implement it later. In this case, we did our best to re-implement their model.

The margin for triplet loss is set to 2 as in the paper~\cite{yang2020hidc}. $\alpha$ and $\beta$ are fixed as 0.5 as in the section 4.2 of the paper, in order to reduce the space of hyper-parameter search. As in the paper, we learn the attribute and object embeddings from scratch, without pretrained GloVe embeddings.
All neural networks are implemented as single MLP layer with ReLU activation. The concept embedding dimension is set to 300.

For the loss terms $\mathcal{L}_{cls}, \mathcal{L}_{conc}, \mathcal{L}_{rec}, \mathcal{L}_{comp}$, we apply different weights $\lambda_{cls}, \lambda_{conc}, \lambda_{rec}, \lambda_{comp}$ to scale them. For MIT-States, we use $\lambda_{cls}=1000$ and $\lambda_{conc}=\lambda_{rec}=\lambda_{comp}=1$, while for UT-Zappos we use $\lambda_{cls}=\lambda_{conc}=\lambda_{rec}=\lambda_{comp}=1$. The optimizer is Adam with learning rate 1e-4 and weight decay 5e-5 for both datasets, except that for UT-Zappos the attribute embeddings are trained with a smaller learning rate of 5e-6 for better performance.

\subsubsection{TCE} For our TCE model, the concept embedding dimension is 256, and we add a single MLP layer to map image features into concept space. We initialize the attribute and object embeddings with 300-dim GloVe~\cite{pennington2014glove} and finetune during training. $g_o$ is implemented as single MLP layer, while $g_a$ contains two MLP layers with 256 hidden units and ReLU activations. $m_o$ is set to 0, $m_c$ is set to 0.5 and $m_r$ is set to 5 for all experiments. $\lambda_{op}=\lambda_{rec}=\lambda_{tri}=1$ for both datasets, $\lambda_{clf}=1000$ for MIT-States, $\lambda_{clf}=1$ for UT-Zappos, and that  $\lambda_{rvc}=0.01$ for both datasets.  We optimize our model using ADAM optimizer with a learning rate $10^{-4}$  for all parameters, except that the attribute embeddings are learned with a learning rate $10^{-5}$.

\end{document}